\documentclass[10pt,letterpaper]{article}

\usepackage{cogsci}

\cogscifinalcopy % Uncomment this line for the final submission 

\usepackage{pslatex}
\usepackage{apacite}
\usepackage{float} % Roger Levy added this and changed figure/table
\usepackage{tcolorbox}
\usepackage{algorithm}
\usepackage{algpseudocode}
\usepackage{amsmath}
\usepackage{algorithmicx}
\usepackage{booktabs}
\usepackage{tabularx}
\usepackage{multirow}
\usepackage{multicol}
\usepackage{amsfonts}
\usepackage{cleveref}
\usepackage{subcaption}
\usepackage[hyphens]{url}
\usepackage[none]{hyphenat} % Sometimes it can be useful to turn off
\title{Computer Vision Models Show Human-Like Sensitivity to Geometric and Topological Concepts}
 
\author{{\large \bf Zekun Wang (zekun@gatech.edu)} \\
  School of Interactive Computing \\
  Atlanta, GA 30332 USA
  \AND {\large \bf Sashank Varma (varma@gatech.edu)} \\
  School of Interactive Computing \\
  School of Psychology \\
  Atlanta, GA 30332 USA}

\begin{document}

\maketitle

\begin{abstract}

With the rapid improvement of machine learning (ML) models, cognitive scientists are increasingly asking about their alignment with how humans think. Here, we ask this question for computer vision models and human sensitivity to geometric and topological (GT) concepts. Under the \emph{core knowledge} account, these concepts are innate and supported by dedicated neural circuitry.
In this work, we investigate an alternative explanation, that GT concepts are learned ``for free'' through everyday interaction with the environment. We do so using computer visions models, which are trained on large image datasets. We build on prior studies to investigate the overall performance and human alignment of three classes of models -- convolutional neural networks (CNNs), transformer-based models, and vision-language models -- on an odd-one-out task testing 43 GT concepts spanning seven classes. Transformer-based models achieve the highest overall accuracy, surpassing that of young children. They also show strong alignment with children's performance, finding the same classes of concepts easy vs. difficult. By contrast, vision-language models underperform their vision-only counterparts and deviate further from human profiles, indicating that naïve multimodality might compromise abstract geometric sensitivity. These findings support the use of computer vision models to evaluate the sufficiency of the learning account for explaining human sensitivity to GT concepts, while also suggesting that integrating linguistic and visual representations might have unpredicted deleterious consequences.

\end{abstract}

\section{Introduction}

Humans learn geometric and topological (GT) concepts through mathematics instruction. Long before they begin formal school, however, they show sensitivity to concepts such as shape, angle, rotation, and translation, leading to the proposal that they are part of \emph{core knowledge}, i.e., innate and supported by dedicated neural circuitry~\cite{Spelke2007}.  This may be the consequence of evolutionary processes building in sensitivity to the mathematical and spatial structure of the world, for example to support navigation~\cite{Gibson1979TheEA,Shepard1994}. Supporting evidence comes from the finding of similar sensitivities in non-human animals~\cite{Chiandetti2007, Pasupathy1999, Vallortigara2018}. The core knowledge account ``explains'' the emergence of GT concepts in young children who have not yet entered school and received formal mathematics instruction.

A contrasting account is that GT concepts are learned ``for free'', through everyday experience in the world. This account has been less studied, perhaps because a strong empirical test would require ``ablation'' studies where organisms are deprived of a typically rich environment in which to learn, raising ethical questions even for animal subjects.

The recent rapid rise of deep learning models provides a new way to test the sufficiency of the learning account.
%% SV: If we need to save space, we can cut Theory of Mind as an example in the next sentence.
These models are increasingly demonstrating human-level performance in domains ranging from language acquisition~\cite{ma2024babysitlanguagemodelscratch,ma2023world,chang2022word} to theory of mind~\cite{ma2024holisticlandscapesituatedtheory,jung2024perceptionsbeliefsexploringprecursory,he2023hitombenchmarkevaluatinghigherorder} to mathematical reasoning~\cite{shah2023humanbehavioralbenchmarkingnumeric,ahn2024largelanguagemodelsmathematical} to concept understanding~\cite{jin2024exploringconceptdepthlarge,vemuri2024deeplearningmodelscapture}. In addition to their fidelity at the behavioral level, there is also increasing evidence of their fidelity at the brain level. Researchers have mapped the layers of convolutional neural network (CNN) models to areas of the human visual system, specifically the ventral visual stream~\cite{he2015deepresiduallearningimage,simonyan2015deepconvolutionalnetworkslargescale,NIPS2012_c399862d,Jacob2021-ek,De_Cesarei2021-gk,Lindsay_2021}.
\citeA{portelance2022neural} argues that despite differences in the underling learning processes of humans and ML models, we can use these models to derive new hypotheses about human cognition, and conversely we can use cognitiv (neuro)science data to train more human-like models. We adopt a similar view.

% In this work, we want to expand the experience or information received by a vision model by using 1) larger model with the Transformer architecture trained on larger dataset as it benefit from scaling up and 2) extra modality such as language to see to what degree the understanding of GT concepts are developed through experiences. We then make comparison to both human and CNN profiles from the previous studies to answer the following research questions. First, with a larger model and more training data, how does the performance of Transformer-based vision models compare to that of CNN-based models and human on GT concepts? Second, how does aligning the text model with the vision model using contrastive loss affect the performance of vision-language models on GT concepts compared to other models and human performance? Third, how well does these neural models align with human profiles on GT concepts? Will a larger, vision-text aligned model result in a better human alignment?

The current study evaluates the sufficiency of the account that GT concepts are learned through experience in the world. It builds on prior studies that have used CNN models and found promising initial results. It replicates their findings and extends their approach to two new and important classes of computer vision models. The first is vision transformer models, which are notable not just for their different architecture but also for their larger size and ability to be trained on more data. The second class is vision-language models, which learn about both the visual-spatial structure of the environment and also its linguistic structure, and the mapping between the two. These models are of potential cognitive interest given that \emph{dual-coding theory}~\cite{Paivio1991} suggests that processing information through both non-verbal and verbal channels should be advantageous. We compare these three classes of ML models to humans in three research questions:

\begin{enumerate}
    \item Do CNNs, transformers, and/or vision-language models approach (or even exceed) the sensitivity to GT concepts shown by children and adults?
    \item Which models / architectures best align with human performance, i.e., find the same classes of concepts easy vs. difficult. Is their alignment close enough to consider them viable cognitive science models?
    \item How does adding language (i.e., symbols) to vision (i.e., space) affect the overall sensitivity and class-by-class alignment of vision-language models?
\end{enumerate}

% \begin{itemize}
%     \item What type of experience help the understanding of GT concepts: larger parameters or extra modality?
%     \item Can ViT outperform CNN in GT concept understand and how it compares with human profile?
%     \item Does modality alignment using contrastive learning give vision model better performance on GT concepts?
%     \item How well does ViT and its corresponding VLM align with human profile?
% \end{itemize}

\section{Related Work}

\subsection{GT concept understanding in humans}

\begin{figure*}
    \centering
    \includegraphics[width=0.85\linewidth]{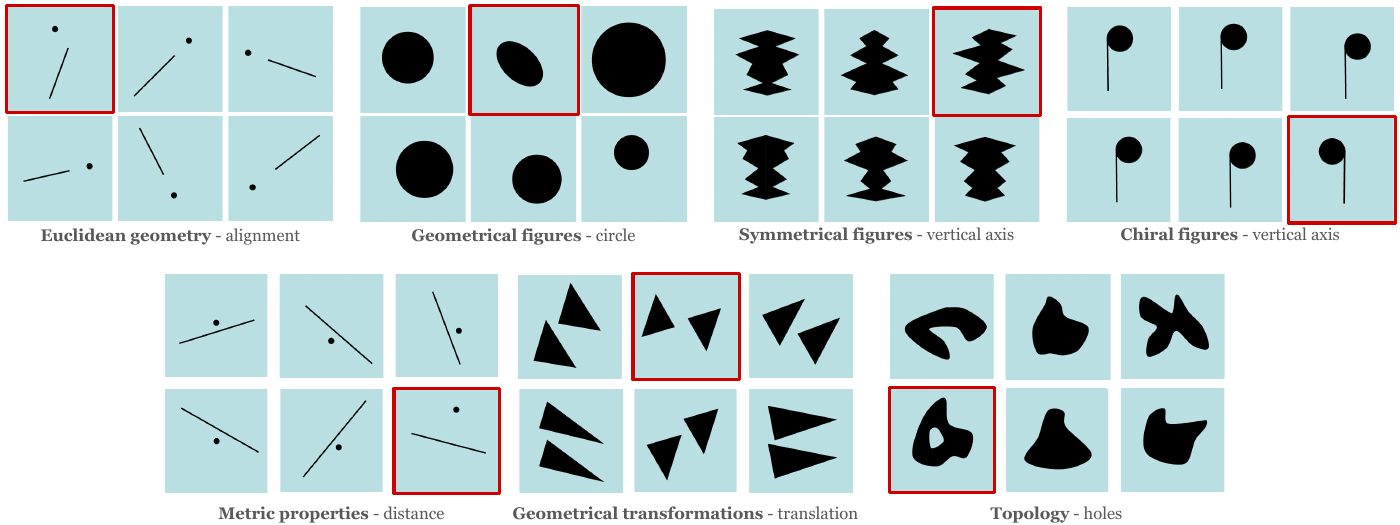}
    \caption{Example stimuli from the 7 categories from~\citeA{Dehaene2006}. The odd-one-out is indicated by the red box.}
    \label{fig:stimuli}
\end{figure*}

\citeA{Dehaene2006} conducted a pioneering and comprehensive study of human sensitivity to GT concepts. They developed a purely visual task that was appropriate even for participants with no formal mathematics education. In this task, each stimulus consists of six images. Five images embody a particular GT concept but the sixth does not. The images otherwise vary in their visual features. The participant is simply asked to indicate the ``odd one out''. They developed stimuli for 43 individual concepts spanning seven broader classes: Topology, Euclidean Geometry, Geometrical Figures, Symmetrical Figures, Chiral Figures, Metric Properties, and Geometrical Transformations.
%%% SV: New figure with examples from all 7 classes looks good!
See \Cref{fig:stimuli} for example stimuli for 7 concepts.

\citeA{Dehaene2006} found that the Mundurucu, an indigenous Amazonian group whose members have no formal schooling, were nevertheless sensitive to 39 of the 43 concepts. Moreover, the Mundurucu adults and children performed comparably. \citeA{Marupudi2023} replicated these findings using a modified 2 alternative forced-choice version of the odd-one-out task, and furthermore showed that cognitive ability in general (i.e., fluid intelligence) and visuospatial ability in particular (i.e., mental rotation) explained only a small portion of the variability in people's performance. \citeA{Izard2009-ok} extended the \citeA{Dehaene2006} study to people from the US. A distinctive contribution was testing both young children (ages 3–6 years) and adults. The young children also displayed sensitivity to GT concepts, displaying above-chance performance for 27 of the 43 concepts. The finding of sensitivity to GT concepts in samples that have had no formal schooling has been generally interpreted as evidence for the core knowledge account.

Interestingly, the three studies found similar variation in the relative difficulty of different GT concepts. For example, participants from all groups -- Mundurucu and Western, children and adults -- were highly accurate for Euclidean Geometry concepts, but found Geometrical Transformations to be difficult. Capturing this variability is an important goal.

\subsection{GT concept understanding in ML models}

Computer vision researchers have documented the sensitivity of CNN models to various GT concepts. \citeA{jaderberg2016spatialtransformernetworks} showed that CNNs with affine transformation are sensitive to parallel lines but insensitive to the affine transformation themselves because affine transformation preserves parallel lines. \citeA{laptev2016tipoolingtransformationinvariantpoolingfeature} demonstrated that training on datasets augmented with rotated images can make CNNs invariant to rotation transformations, which potentially \emph{works against} sensitivity to the rotation concept within the Geometric Transformation class. \citeA{Mumuni2021} provide a comprehensive review of approaches to extending the CNN architecture to handle non-trivial geometric transformations. 

The goal of this research -- to develop models that are invariant to different geometric transformations, for the purpose of better generalization -- makes sense for computer vision. For cognitive science, the question is whether these models learn human-like sensitivities. \citeA{hsu2022geoclideanfewshotgeneralizationeuclidean} tested the sensitivity of CNNs to Euclidean geometry concepts, finding that humans outperform CNNs pre-trained on the ImageNet~\cite{imagenet}.
In the direct precursor to the current study, \citeA{upadhyay2025alignment} investigated the sensitivity of a broad range of CNN models to the 43 GT concepts of \citeA{Dehaene2006}. ResNet-18 \cite{he2015deepresiduallearningimage} showed the highest overall accuracy, though it was still below that of the young children of the \citeA{Izard2009-ok} study. They also found medium-size correlations ($r \approx 0.55$) between the CNN models and humans in their average accuracies across the seven classes of GT concepts. Thus, there is some evidence of alignment, but also much room for improvement.
In another related study, \citeA{campbell2024humanlikegeometricabstractionlarge} explored the sensitivity of vision transformer models such as DINOv2~\cite{oquab2024dinov2learningrobustvisual} and CLIP~\cite{radford2021learningtransferablevisualmodels} to the Geometric Regularity stimuli of \citeA{sablemeyer2022language} and the Geometric Parts and Relations stimuli of \citeA{hsu2022geoclideanfewshotgeneralizationeuclidean}. These newer models showed stronger alignment with human performance than the older CNN model ResNet-50 \cite{he2015deepresiduallearningimage}.

% In our work, we study the sensitivity to GT concepts on a variety of CNN- and Transformer-based models and compare them to the human performance from previous cognitive science studies.

% For instance, in mathematical cognition, \todo{[need citation for the AAAI paper]} shows the sensitive to GT concepts in CNN models that GT concepts are likely acquired through the experience and training data, and~\cite{campbell2024humanlikegeometricabstractionlarge} demonstrates a better human alignment in geometric abstraction tasks with larger pre-trained neural networks.

\section{Method}

\subsection{Models}

%% SV: I find the "TVM" acronym jarring. Can we just call this class of models "Transformers"? I know this is a little problematic because of the CLIP (ViT) model....
The current study explored the sensitivity to GT concepts and the human alignment of three classes of models -- CNNs, Transformers, and Vision-Language Models (VLMs). We selected the following examples of each class:

\begin{itemize}
    \item \textbf{CNNs}: ResNet-50, ResNet-18~\cite{he2015deepresiduallearningimage}, and EfficientNet~\cite{tan2020efficientnetrethinkingmodelscaling}
    \item \textbf{Transformers}: Vision Transformer (ViT)~\cite{dosovitskiy2021imageworth16x16words} and DINOv2~\cite{oquab2024dinov2learningrobustvisual}
    \item \textbf{VLMs}: CLIP~\cite{radford2021learningtransferablevisualmodels} with either a CNN or ViT  vision backbone, and ALIGN~\cite{jia2021scalingvisualvisionlanguagerepresentation}
\end{itemize}

\noindent Some models were selected based on prior research. In particular, \citeA{upadhyay2025alignment} found ResNet-18 to be the CNN model that showed the greatest sensitivity to the GT concepts of \citeA{Dehaene2006}, and \citeA{campbell2024humanlikegeometricabstractionlarge} explored the geometric sensitivity of CLIP, DINOv2, and ResNet-50.

We use versions of these models that are publicly available on \textit{Huggingface}. For a fair comparison to the CNN-based models, we used the ``base'' size variants of the Transformer and VLM models. 
% All vision-only models (Transformers and CVMs) are trained on the image classification tasks where the training data consists of (image, class label) pairs. 
All vision-only models (CNNs and Transformers) except for DINOv2 were trained on image classification tasks using supervised (image, class label) pairs.
DINOv2 was trained in a self-supervised manner on the LVD-142M dataset~\cite{oquab2024dinov2learningrobustvisual}, a large-scale data set curated specifically for this model. It contains ImageNet-21k without requiring manual labels for its learning objective.
Specifically, it uses a self-distillation loss on both image-level and patch-level representations, where a teacher model provides soft targets for a student model to learn robust representations.
All CNN models were trained on ImageNet-1k, a subset of the much larger ImageNet-21k dataset used to train the Transformers.
The VLM models were trained on the contrastive loss between (image, caption) pairs to maximize the similarity between the hidden representations of related images and captions. 
The training datasets for contrastive learning usually have to be much larger in size than those for learning image classification, and are not publicly available.
Notably, the publicly available version of ALIGN was trained on a publicly available dataset, COYO~\cite{kakaobrain2022coyo-700m}, and achieves comparable or superior performance to the original, non-public ALIGN model. \Cref{tab:model_comparison} summarizes the key properties of the models and their training.

\subsection{Stimuli and Datasets}
We use the stimulus set from \citeA{Dehaene2006}. The set consists of 43 stimuli (or `tasks'). Each stimulus corresponds to a GT concept (e.g., parallel lines). It is composed of 6 images, 5 of which embody that concept and 1 of which does not. The task is to select the odd-one-out image (see \Cref{fig:stimuli}). The 43 stimuli can be grouped into 7 categories (\emph{N}): Topology (4), Euclidean Geometry (8), Geometrical Figures (9), Symmetrical Figures (3), Chiral Figures (4), Metric Properties (7), and Geometrical Transformations (8). 

We obtained three human datasets from published studies. The first dataset comes from \citeA{Dehaene2006}, who measured the performance of Mundurucu adults and children. This is an indigenous Amazonian group whose members inhabit isolated communities and receive little or no formal education. (That study found comparable performance in the two age groups.) From \citeA{Izard2009-ok} we have the performance of Western children between the ages of 3 and 6, forming the second dataset, and from Western adults between the ages 18 and 25, forming the third dataset.

\begin{table}[ht]
\centering
\scriptsize
% \small
\begin{tabularx}{\columnwidth}{p{1.25cm}|p{1.3cm}p{1cm}p{1.77cm}p{1.2cm}}
\toprule
\textbf{Type} & \textbf{Model} & \textbf{\# Params} & \textbf{Dataset} & \textbf{\# Training Data} \\
\midrule
\multirow{2}{*}{Transformers} & ViT & 86M & ImageNet-21k & 14M \\
                     & DINOv2 & 86M & LVD-142M & 142M \\
\midrule
\multirow{3}{*}{CNN} & ResNet-50 & 26M & ImageNet-1k & 1.3M \\
                     & ResNet-18 & 11M & ImageNet-1k & 1.3M \\
                     & EfficientNet & 66M & ImageNet-1k & 1.3M \\
\midrule
\multirow{2}{*}{VLM} & CLIP-RN50 & 90M & WIT & 400M \\
                     & CLIP-ViT & 150M & WIT & 400M \\
                     & ALIGN & 174M & COYO & 700M \\
\bottomrule
\end{tabularx}
\caption{The eight models, their parameters, datasets, and training data sizes.}
\label{tab:model_comparison}
\end{table}

\subsection{Evaluation on Neural Models}

We followed the evaluation method of the earlier CNN study by~\citeA{upadhyay2025alignment}; see also \citeA{campbell2024humanlikegeometricabstractionlarge}, \citeA{muttenthaler2023human}, and \citeA{muttenthaler2023improving}. We first re-scaled and cropped the 6 images in each stimulus to a size of $224 \times 224$. We then passed the images into the pre-trained models and collected their representations from the final hidden layer \footnote{The layer before any projection or classification.}. In the VLMs, the images were processed entirely by the image encoders. After collecting the image representations, we compute pairwise cosine similarities between the 6 images, and then for each image we computed its average cosine similarity to the other 5 images. The image with the lowest average cosine similarity was selected as the odd-one-out. We repeated this process for all stimuli and for all models.

\section{Results}
% \TODO{still refining the aesthetics of the plots}

\subsection{Overall sensitivity}

\begin{figure}[ht]
    \centering
    \includegraphics[width=0.85\linewidth]{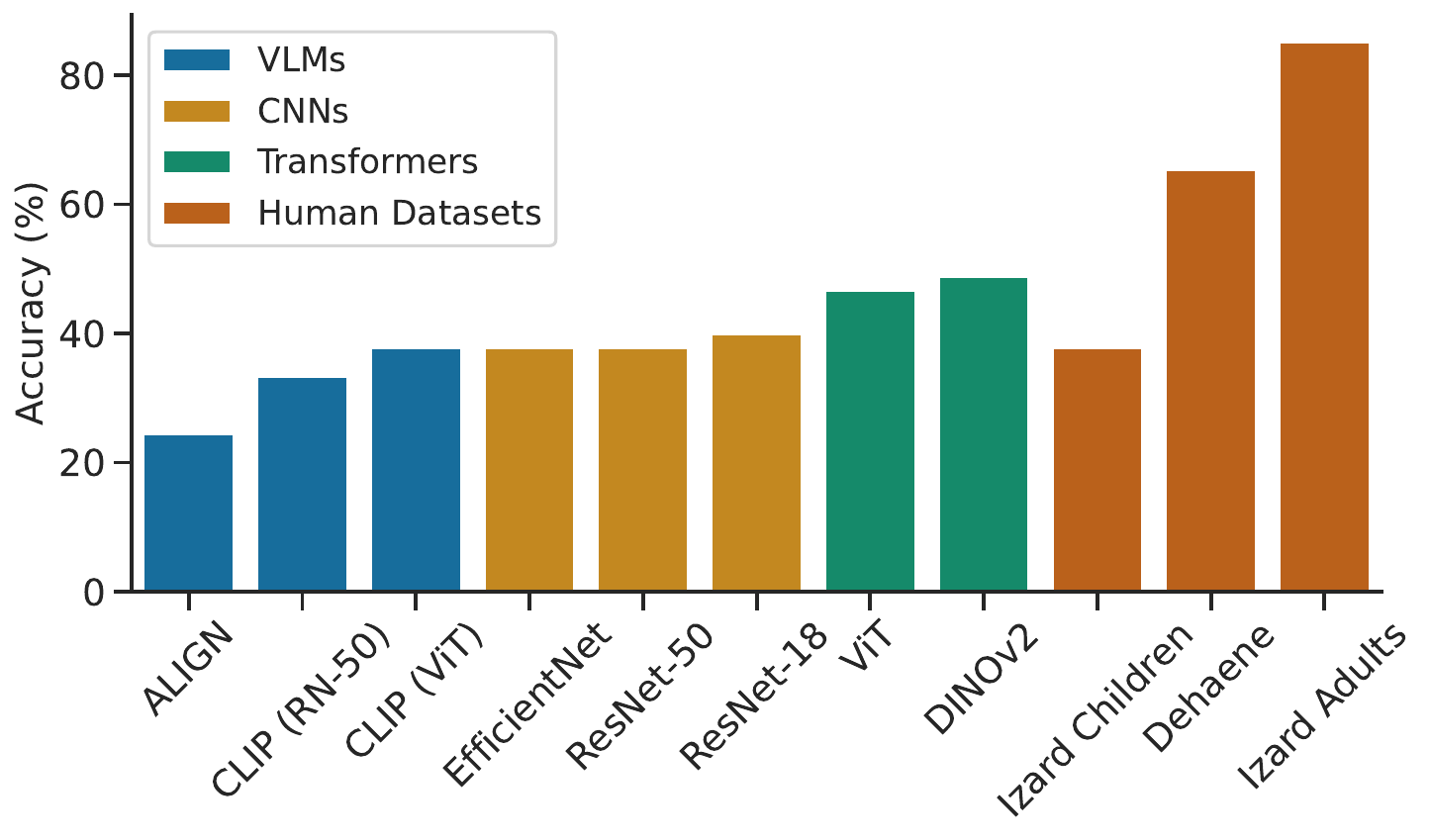}
    \caption{Average accuracy on the odd-one-out task for both the ML models and the human participants.}
    \label{fig:overall-accu}
\end{figure}

For each model, we computed the overall accuracy averaged across the 43 concepts. \Cref{fig:overall-accu} shows these measures along with the overall accuracy in each of the human datasets. All but one of the models achieve higher overall accuracies than would be predicted by chance ($16.67\%$), $z$s $> 2.3$, $p$s $< .02$; the only exception is ALIGN ($z = 1.187, p = .118$).

%%% Technically, to make the statement in the next sentence, you want to statisically compare each of the average model accuracies to .167. This is a z-test of proportions (https://online.stat.psu.edu/statprogram/reviews/statistical-concepts/proportions). If you give the z and p values for the 8 tests corrsponding to each of the models, I will revise the next sentence to properly report this information.

% alpha = 0.05
% null: mean = 0.1667
% ALIGN:            z_stat=1.187,   p_val=0.118, Fail to reject
% CLIP (RN-50):     z_stat=2.318,   p_val=0.01,  Reject null
% CLIP (ViT):       z_stat=2.855,   p_val=0.002, Reject null
% EfficientNet:     z_stat=2.855,   p_val=0.002, Reject null
% ResNet-50:        z_stat=2.855,   p_val=0.002, Reject null
% ResNet-18:        z_stat=3.123,   p_val=0.001, Reject null
% ViT:              z_stat=3.943,   p_val=0.0,   Reject null
% DINOv2:           z_stat=4.227,   p_val=0.0,   Reject null
% Izard Children:   z_stat=2.849,   p_val=0.002, Reject null
% Dehaene:          z_stat=6.709,   p_val=0.0,   Reject null
% Izard Adults:     z_stat=12.606,  p_val=0.0,   Reject null

The Transformers outperform both the CNNs and VLMs, achieving 46.67\% (ViT) and 48.89\% (DINOv2). By comparison, the best-performing CNN, ResNet-18, achieves an accuracy of 40\%, and the best-performing VLM, CLIP (ViT), achieves an accuracy of 37.78\%.

Notably, only the Transformers show higher overall accuracies than the 3-6 year old children of ~\citeA{Izard2009-ok} (37.72\%). However, their performance is still far below the 65.34\% accuracy of the Mundurucu adults and children ~\cite{Dehaene2006} and the 85.10\% accuracy of the Western adults~\cite{Izard2009-ok}. 

The CNNs show comparable overall accuracies to the 3-6 year old children of ~\citeA{Izard2009-ok}. Interestingly, the two deeper models (EfficientNet and ResNet-50) show slightly lower sensitivity to GT concepts (both 37.78\%) than the shallower model (ResNet-18, 40\%), paralleling the findings of~\citeA{upadhyay2025alignment}.

The VLMs show the least sensitivity to GT concepts. The best VLM, CLIP (ViT), achieves an overall accuracy of 37.78\%, while the ResNet-50 variant of CLIP, CLIP (RN-50), achieves an overall accuracy of only 33.33\%. ALIGN is the worst-performing model, achieving an overall accuracy of only 24.44\% despite being trained on the largest dataset.

\begin{figure*}[ht]
    \centering
    \includegraphics[width=0.9\linewidth]{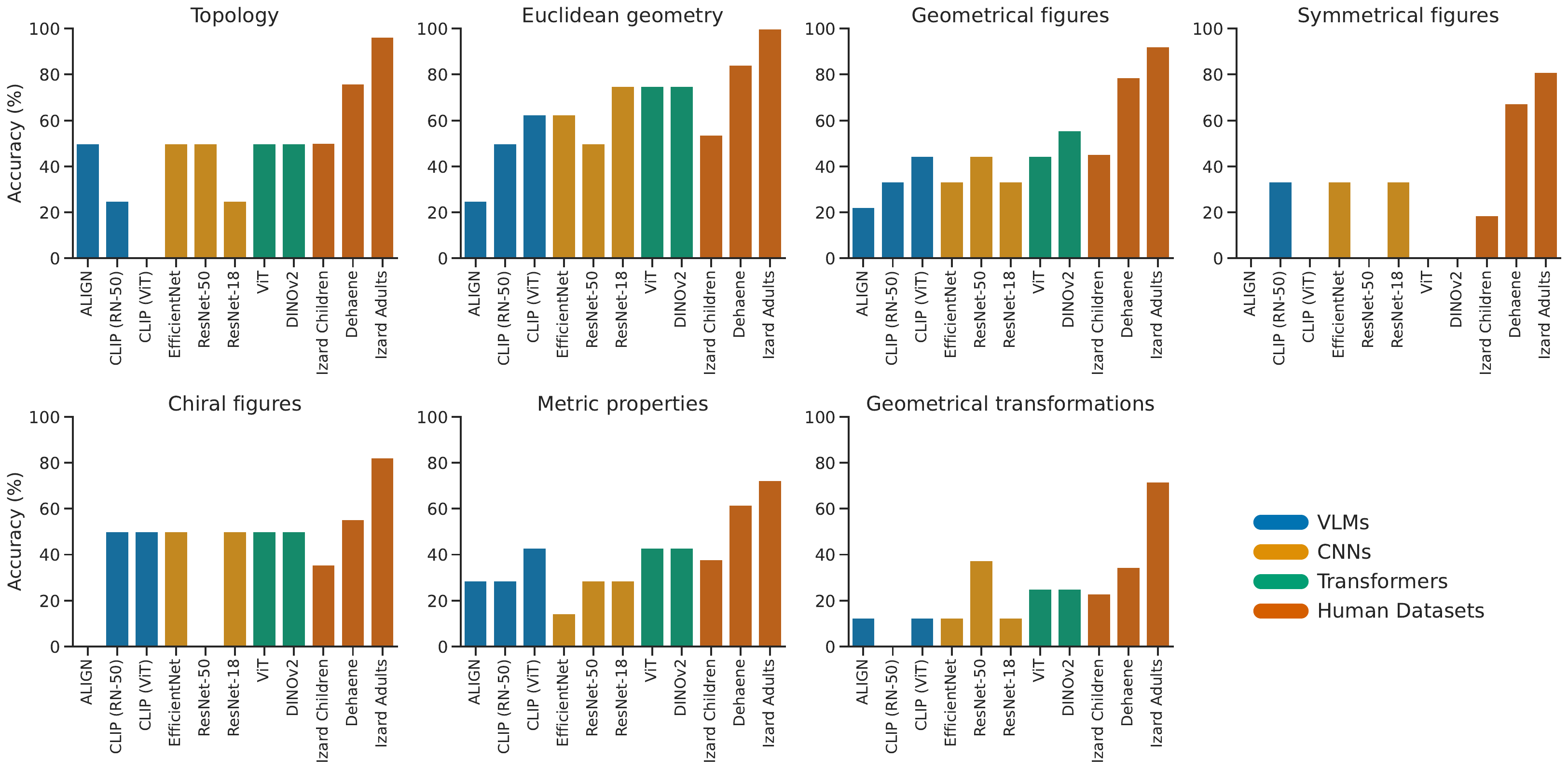}
    \caption{Accuracy profiles of the models and humans for each of the 7 classes of GT concepts.\vspace{-10px}}
    \label{fig:by-task}
\end{figure*}

%% SV: This figure is great. (Though it is missing 2 models!) However, it is just showing the correlations that are in the next Figure in a different form. It can be deleted to save space.
% \begin{figure}[th]
%     \centering
%     \includegraphics[width=1\linewidth]{full_paper//pics/odd-one-out-accuracy-by-task-by-model.pdf}
%     \caption{Accuracy profiles grouped by VLMs and their corresponding vision-only models and human participants data in the odd-one-out tasks. Specifically, CLIP (ViT) uses ViT as the vision backbone, CLIP (RN-50) uses ResNet-50, and ALIGN uses EfficientNet.\vspace{-10pt}}
%     \label{fig:by-vlm}
% \end{figure}

\subsection{Sensitivity by class}

To further probe the sensitivities of the models, we evaluated their average accuracy for each of the 7 classes; see \Cref{fig:by-task}. A number of notable patterns are visible at this finer-grain level of analysis.

First, the high overall accuracy of the Transformers is not driven by superior performance on one or two classes. Rather, ViT and DINOv2 outperform all of the other models on 6 of the 7 classes, with a few exceptions: The Transformers perform poorly on Symmetrical Figures concepts, and the CNN model ResNet-50 achieves a higher accuracy for the Geometric Transformations class.

Second, the Transformers do not just achieve higher overall accuracies than the 3-6 year old children of \citeA{Izard2009-ok} (see \Cref{fig:overall-accu}). They are as accurate or more accurate than the children on all 7 of the classes. Conversely, they are less accurate than the two samples that include adult participants on all 7 of the classes.

Third, the Euclidean Geometry concepts are the easiest, with both models and humans achieving their highest accuracies for this class. Conversely, the Geometrical Transformations and Symmetrical Figures concepts are the most difficult, with especially the models achieving their lowest accuracies for these classes. The biggest discrepancy between the models and humans is on Symmetrical Figures, for which no model achieves greater than 33\% accuracy. Also notable is that even the Transformers, the best-performing of all models, struggle with Symmetrical Figures. However, it must be noted that this class contains the fewest concepts, which make the data patterns here somewhat tentative.

\subsection{Alignment to human profiles}
\begin{figure*}[ht]
    \centering
    \includegraphics[width=0.85\linewidth]{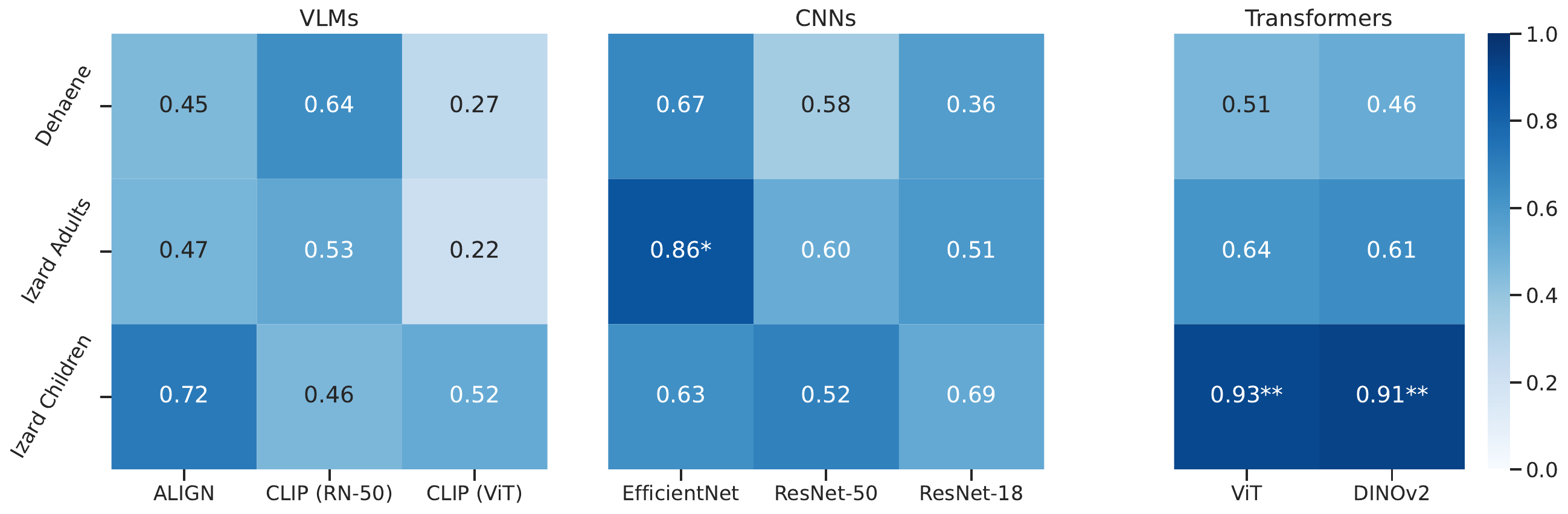}
    \caption{Heatmap of Pearson $r$ coefficients between human participants’ sensitivity profiles and the models’ sensitivity profiles across the 7 classes of GT concepts. Note that * denotes $p < 0.05$ and ** denotes $p < 0.01$. \vspace{-10px}}
    \label{fig:human-corr}
\end{figure*}

That a model is highly accurate across the concepts of a class is not useful if, in fact, humans find that class to be difficult, given that our goal is to find alignment between model and human performance. To most directly evaluate this alignment, we compute the Pearson correlation ($r$) between the profile of accuracies across the 7 classes for each of the 8 models and the 3 human datasets; see \Cref{fig:human-corr}. There are a number of patterns to notice.

%% SV: For each correlation, it would be good to know if it is statistically different than 0. Do these tests. Then put an "*" besides each such correlation Finally, add to the Figure caption: "Note that * denotes $p < .05$".

First, the class accuracy profiles of the Transformers achieve the closest alignment to those of the human datasets. This is particularly striking for the GT sensitivities of the 3-6 year old children in the \citeA{Izard2009-ok} study ($r$s $> 0.90$). Thus, the Transformers offer the best account of the human data at all three levels: overall accuracy, by-class accuracy, and by-class accuracy profile (i.e., human alignment).

Second, among the other models, the CNN model EfficientNet also achieves close alignment with the human datasets, particularly the adults in the \citeA{Izard2009-ok} study ($r = 0.86$). This raises the question of whether there is continuity in the development of GT concepts, i.e., given a model architecture, more parameters results in better alignment with children's performance, giving way to better alignment with adult performance. Or whether there is a qualitative shift across development, that is, a shift from a transformer-like architecture to a CNN-like architecture.

Third, the CLIP (ViT) model shows the worst alignment to the class accuracy profiles of adults, and the CLIP (RN5-50) model shows the worst alignment to the 3-6 year old children in the~\citeA{Izard2009-ok} study.

Fourth, among the datasets, the ML models generally achieve closer alignment to the children's data of \citeA{Izard2009-ok} than to the other two datasets, both of which contain adult data.

\section{Discussion}

%% SV: I moved this text down from the Introduction and turned it into a nice first paragraph for the Discussion.
Our results find that the Transformers show more promise as a cognitive science models of geometric and topological concepts than either CNNs or VLMs. At a coarse-grain level, they show the highest overall accuracies -- higher even than the 3-6 year old children from the \citeA{Izard2009-ok} study; see \Cref{fig:overall-accu}. At the finer-grain level of the 7 classes, their accuracy profiles most closely align with that of the 3-6 year old children from the \citeA{Izard2009-ok} study; see \Cref{fig:by-task} and \Cref{fig:human-corr}.
An unexpected finding is that the VLMs trained with contrastive loss give both the worst overall and class-level performance, and also show the worst alignment to human profiles. This is surprising conceptually given the findings that support dual-coding theory~\cite{Paivio1991}. This is also surprising technically: the sizes of the VLM models and the datasets on which they are trained are much larger than is the case for the Transformers and CNNs. This may suggest that current modality alignment techniques, such as maximizing the similarity between the hidden representation of both textual and visual information in a shared vector space, fail to fuse the two streams of information in a way which builds sensitivity to GT concepts, and that more human-like alignment methods are needed.

\subsection{Transformer models show higher sensitivity to GT concepts}

\Cref{fig:overall-accu,fig:by-task} demonstrate that the Transformers, which have more parameters than CNNs and are trained on a larger dataset (e.g., ImageNet-21k with 14 million training samples compared to ImageNet-1k with 1.3 million training samples used for CNNs), exhibit higher sensitivity to GT concepts, and even surpass the overall accuracy of the 3-6 year old children from the \citeA{Izard2009-ok} study.
These findings support the claim by~\citeA{upadhyay2025alignment} that GT concepts can be learned ``for free'' as a consequence of training on more generic visual processing tasks such as image classification.
%%% SV: I don't think these citations are needed here.
% ~\cite{Gelman2009,Walsh2016}. 
This contrasts with the ``core knowledge'' view that humans possess innate knowledge of mathematical concepts which requires minimal external input to activate \cite{Spelke2007}. 

Transformers are generally larger than CNNs and trained on more data. Beyond this, we hypothesize that Transformers show higher sensitivity to GT concepts than CNNs for two reason. First, Transformers relax the constraint of transformation-invariants~\cite{raghu2022visiontransformerslikeconvolutional}. Second, their self-attention mechanism may learn both global and local object geometry whereas CNNs cannot~\cite{cordonnier2020relationshipselfattentionconvolutionallayers}; the result is a more robust representation than CNNs. 
As a result, Transformers can acquire the sensitivity to object transformations and geometric/topological concepts.
A direction for future research is to analyze the salience maps to each GT concept stimulus in Transformers and CNNs, to better understand the aspects that drive their sensitivities to GT concepts. 
For instance, one could collect the attention weights for each layer of a Transformer model and directly use them as the salience map. This is because the attention score tells the importance of current image patch with respect to other patches when making classification decisions. 
For CNNs, which do not possess attention mechanisms, we can compute the gradients of the target class score with respect to the feature maps of a convolutional layer, weighting these feature maps by the averaged gradients, and then summing them up~\cite{Selvaraju_2019}. A higher gradient score means that a small change in the feature map would cause a significant change in the output, indicating that the feature map is highly relevant to the model’s performance.

% A future research direction is to analyze salience maps for GT concept stimuli in Transformers and CNNs, illuminating the visual features driving their respective sensitivities. For Transformers, attention weights from each layer, which indicate inter-patch importance for classification, can directly serve as salience maps. For CNNs, which lack attention mechanisms, gradient-based methods like Grad-CAM~\cite{Selvaraju_2019}can compute the importance of feature maps for the model's decisions by analyzing class score gradients

%%% SV: I confess I don't have the background to really understand/challenge any points in this section. I made some edits but will largely have to trust your knowledge here.
\subsection{Modality alignment with text reduces the sensitivity to GT concepts}

Humans process information through both verbal and non-verbal channels. ~\citeA{Paivio1991} argues that these channels work independently and interactively to enhance understanding and memory, and enable more robust learning because the information encoded in both formats provides multiple pathways for retrieval.
Recent research in multi-modal alignment indeed is consistent with this view. For example, CLIP~\cite{radford2021learningtransferablevisualmodels} significantly outperforms CNNs in image retrieval and classification tasks. It is therefore surprising, perhaps, that the CLIP models show lower sensitivity to GT concepts and worse alignment to human accuracy profiles compared to CNNs and Transformers. 
We argue that the effectiveness of current VLMs is limited by their core alignment strategy: optimizing similarity between image and caption representations. This approach struggles with inputs where the visual content is hard to describe naturally in text, such as the abstract synthetic images found in odd-one-out tasks. In these cases, VLMs fail to build robust representations that use both vision and text.
Our findings (\Cref{fig:by-task}) support this, showing reduced VLM sensitivity to abstract concepts like those of Topology and Geometrical Transformation compared to vision-only models – categories that are difficult to map textually onto specific pixels. In contrast, VLM performance on Metric Properties and Geometric Figures is less impacted, as descriptions involving distance, relative position, or basic shapes map more easily to pixel-level information.
\citeA{conwell2022testingrelationalunderstandingtextguided} also argues that failures in image generation guided by CLIP embeddings reveal a lack of a true compositional understanding of textual relations. Their results suggest that these models often reproduce statistically common image-caption pairings from training data rather than accurately constructing the specific relationship requested in the prompt, suggesting that CLIP models fail to capture representations for out-of-distribution text and image.

\subsection{From cognitive alignment to developmental alignment}

Transformers exceed the overall accuracy of the 3-6 year old children of the \citeA{Izard2009-ok} study (see \Cref{fig:overall-accu}), and their performance profiles across the 7 classes of GT concepts are strongly correlated with those of the children (see \Cref{fig:human-corr}). In both respects, they exceed the cognitive alignment of the other evaluated models
%...suggesting that these models can serve as \emph{“animal models”} of child-like visual cognition. 
%By studying Transformers, cognitive scientists can potentially gain insights into how children learn and represent geometric information over developmental time.
A key question is whether Transformer models show not just strong cognitive alignment but also strong developmental alignment. A key advantage of computational models over human participants is their inspect-ability and manipulate-ability. Future research can measure their sensitivity to GT concepts over training and evaluate whether this tracks the growing sensitivity shown by children over development. For example, we can track when during training a Transformer model becomes sensitive to shapes before transformations, and compare this ordering of unfolding sensitivities to that of the developing child. At a finer-grain level, we can analyze how representations of object parts, boundaries, or spatial relationships might evolve in the “developing” parameters of Transformers. 

We can also conduct experiments on Transformers to test the causality of the learning account. A controlled, long-term ablation study in human participants -- systematically withholding certain types of stimuli or training at particular times during development -- would be both unethical and practically infeasible. By contrast, it is straightforward to perform parallel computational studies of computer vision models or study their development over different training ``curricula''. Such studies will potentially lead to novel hypotheses about \emph{when} and \emph{how} young children acquire specific cognitive competencies~\cite{evanson2023languageacquisitionchildrenlanguage,ma2024babysitlanguagemodelscratch}. 
%For example, we can track at which training epoch a Transformers becomes sensitive to shapes versus transformations, and then compare that trajectory to existing child behavioral data.
%%% SV: I moved this point up, leaving this sentence dangling, so I deleted it.
% One immediate future work is to study the developmental trajectory of sensitivity to GT concepts in Transformers. 
%Because most Transformers are originally trained on classification objectives (e.g., predicting an ImageNet label), another line of future work is to interrogate whether humans also learn to perceive objects by maximizing predictive power. If so, this might support theories that conceptual development is grounded in increasingly refined predictions about sensory experience. Investigating how humans and Transformers both generalize to new classes of geometric transformations—or systematically fail under certain conditions—can reveal gaps in current learning paradigms and inspire a more human-like learning algorithms.
% Overall, our results provide compelling evidence for using model checkpoints during training to study the development of GT concepts. While Transformers are not perfect analogues of the human visual system...
The strong alignment especially between the Transformer models and and the young children in their sensitivity to GT concepts suggests their potential for cognitive and developmental science.

\bibliographystyle{apacite}

\setlength{\bibleftmargin}{.125in}
\setlength{\bibindent}{-\bibleftmargin}

\bibliography{cameraready}

\end{document}